\documentclass[times, review, 10pt]{elsarticle}

\usepackage{amssymb}
\usepackage{amsmath}
\usepackage{graphicx}
\usepackage{color}
\usepackage{framed}
\usepackage{adjustbox}
\usepackage{booktabs}
\definecolor{shadecolor}{rgb}{0,1,0}
\usepackage{multirow}
\usepackage{xspace}
\usepackage{colortbl} 
\usepackage{xcolor}
\usepackage{tabularx}
\usepackage{subcaption}
\usepackage{rotating}

\newcommand{\etal}{\textit{et al.}\xspace}

\newcommand{\etc}{\textit{etc.}\xspace}

\usepackage{hyperref}
\hypersetup{
    colorlinks=true, %
    linkcolor=blue,  %
    filecolor=blue,  %
    urlcolor=blue,   %
    citecolor=blue,  %
}

\journal{Pattern Recognition}

\begin{document}
	
	\begin{frontmatter}
		
		\title{Coarse-to-Fine Crack Cue for Robust Crack Detection}

		\author[label1,label2,label3]{Zelong Liu}
        \ead{zelongliu@whu.edu.cn}
		\author[label1,label2,label3]{Yuliang Gu}
        \ead{yuliang_gu@whu.edu.cn}
        \author[label1,label2,label3]{Zhichao Sun}
        \ead{zhichaosun@whu.edu.cn}
        \author[label1,label2,label3]{Huachao Zhu}
        \ead{huachao.zhu@whu.edu.cn}
        \author[label1,label2,label3]{Xin Xiao}
        \ead{xinxiao@whu.edu.cn}
        \author[label1,label2,label3]{Bo Du}
        \ead{dubo@whu.edu.cn}
        \author[label4]{Laurent Najman}
        \ead{laurent.najman@esiee.fr}
		\author[label1,label2,label3]{Yongchao Xu \corref{cor1}}
        \cortext[cor1]{Corresponding author}
        \ead{yongchao.xu@whu.edu.cn}

		\address[label1]{
            National Engineering Research Center for Multimedia Software, Wuhan University, Wuhan,China }
        \address[label2]{
            Institute of Artificial Intelligence, School of Computer Science, Wuhan University,Wuhan,China}
        \address[label3]{Hubei Key Laboratory of Multimedia and Network Communication Engineering, Wuhan University,Wuhan,China }
        \address[label4]{Université Gustave Eiffel, CNRS, Laboratoire d'Informatique Gaspard-Monge (LIGM), Marne-la-Vall, France}

\begin{abstract}
Crack detection is an important task in computer vision.  Despite impressive in-dataset performance, deep learning-based methods still struggle in generalizing to unseen domains. The thin structure property of cracks is usually overlooked by previous methods. In this work, we introduce CrackCue, a novel method for robust crack detection based on coarse-to-fine crack cue generation. The core concept lies on leveraging the thin structure property to generate a robust crack cue, guiding the crack detection. Specifically, we first employ a simple max-pooling and upsampling operation on the crack image. This results in a coarse crack-free background, based on which a fine crack-free background can be obtained via a reconstruction network. The difference between the original image and fine crack-free background provides a fine crack cue. This fine cue embeds robust crack prior information which is unaffected by complex backgrounds, shadow, and varied lighting. As a plug-and-play method, we incorporate the proposed CrackCue into three advanced crack detection networks. Extensive experimental results demonstrate that the proposed CrackCue significantly improves the generalization ability and robustness of the baseline methods. The source code will be publicly available. 
\end{abstract}

		
	\end{frontmatter}

\section{Introduction}
\label{sec:intro}

Cracks are one of the most common defects on the surfaces of many public facilities, such as concrete pavements, bridges, and tunnel ceilings. If these cracks are not detected and maintained early, they can lead to continuous deterioration of the surface structure, reducing the facility's lifespan and causing serious traffic and safety hazards. Therefore, detecting surface cracks~\cite{chen2024mind} at an early stage is crucial.
Factors in practice, such as complex backgrounds and diverse environmental elements, desire higher requirements for the robustness and generalization of the detection model. Traditional machine learning based methods CrackTree~\cite{zou2012cracktree} and CrackForest~\cite{Crackforest}, as well as edge detection based methods like HED~\cite{HED} and RCF~\cite{rcf} are initially employed for crack detection. However, they often lack in semantic understanding, which is crucial for distinguishing cracks from other similar patterns in complex environments. Recent works have shifted towards leveraging advanced semantic segmentation methods to amplify the semantic discrimination. For instance, DeepCrack~\cite{deepcrack} employs multi-scale features to enhance the detail and accuracy of crack representation. CrackVit~\cite{quan2023crackvit} incorporates transformer architecture, benefiting from their ability to process global context information for crack. Similarly, CrackFormer~\cite{liu2021crackformer} integrates transformer-based network and introduces new scaling-attention module to suppress non-crack features. While these methods have achieved impressive in-dataset accuracy, they often struggle to generalize well across different datasets with varying backgrounds, textures, and lighting conditions, making them difficult to handle complex and dynamic real-world scenarios.

In this work, we propose CrackCue, a novel method for robust crack detection based on coarse-to-fine crack cue generation. The key concept is that we introduce a crack cue, which is concatenated with the original image before feeding into the crack detection network. The crack cue incorporates prior information of the crack regions. Specifically, based on the inherent characteristics of crack structure (being thin and usually darker than the context), we employ max-pooling and upsampling operation to obtain a coarse crack-free background image. The coarse crack cue can be obtained by the absolute difference between the original image and this coarse background image. We then proceed to obtain the robust fine crack cue given by the difference of the original image and finely reconstructed background image with a reconstruction network from the coarse background image. This step ensures that the fine cue remains effective and consistent in varying and challenging image conditions, including complex backgrounds, shadow and varied lighting. 

\begin{figure}[t]
  \centering
  \includegraphics[width=\linewidth]{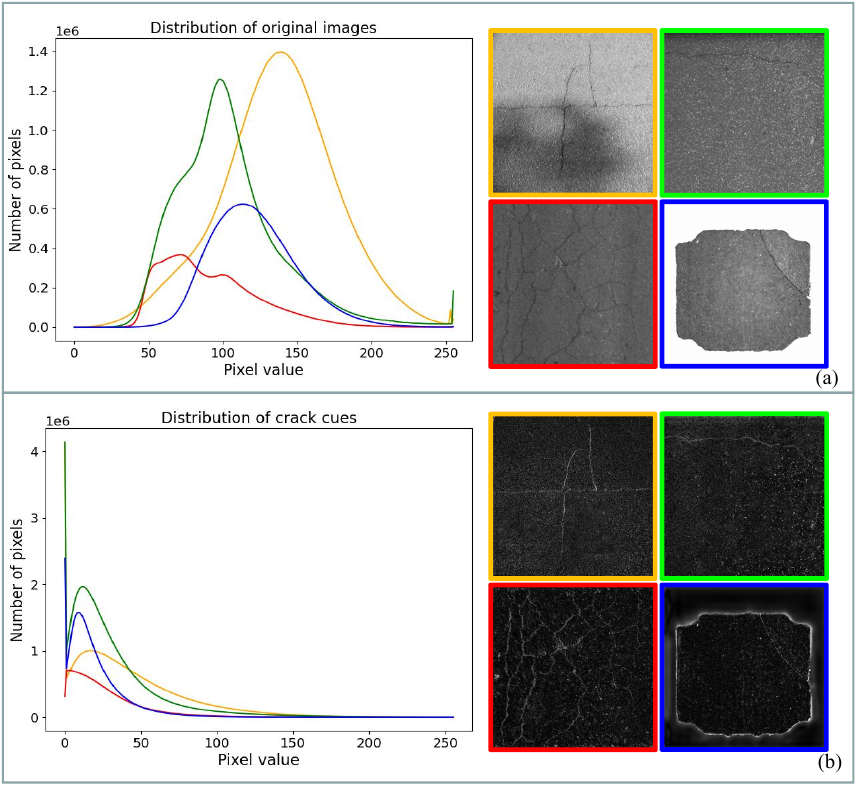}
  \caption{Comparison of distribution of pixel intensity in (a) original image and (b) crack cue on four crack datasets. Orange: CrackTree260; Green: CrackLS315; Red: CRKWH100; Blue: Stone331. The domain gaps between crack cues are significantly reduced compared with the gaps on original images.}
  \label{fig:single}
\end{figure}

Figure~\ref{fig:single} presents an analysis of pixel intensity distributions of the original images and the generated crack cues in four widely-used crack datasets: CrackTree260~\cite{zou2012cracktree}, CrackLS315~\cite{deepcrack}, CRKWH100~\cite{deepcrack}, and Stone331~\cite{deepcrack}. The differences in pixel intensity distributions of original images across four datasets highlight a substantial domain gap. However, the crack cues exhibit roughly similar distributions, reducing the gap between these four domains. This is because the original and reconstructed background images are similarly affected by factors such as image contrast and background texture changes. Consequently, their absolute difference that forms the crack cue effectively offsets the domain gap, resulting in a more robust crack cue representation.

To verify the effectiveness of the proposed CrackCue, we conduct cross-dataset evaluations on four crack datasets. Extensive experiments indicate that the proposed CrackCue significantly improves the generalization ability and robustness of some popular baseline methods for crack detection. Extension to retinal blood vessel segmentation further demonstrates that the proposed CrackCue is not only effective for crack images but also for other curvilinear structural images. The main contributions of this paper are summarized as follows:
\begin{itemize}
\item We introduce a simple yet novel plug-and-play crack cue concept to guide crack detection. The crack cue provides robust crack information that is unaffected by complex backgrounds, shadows, and uneven illumination, thereby enhancing the generalization ability of crack detection networks.

\item We propose a simple yet effective coarse-to-fine generation process for robust crack cue, further boosting the performance.

\item Extensive experiments demonstrate that our CrackCue significantly improves the generalization ability and robustness of some state-of-the-art methods.
\end{itemize}

\section{Related work}  \label{sec:related work}

\subsection{Traditional crack detection}

Under normal lighting conditions, cracks typically appear darker than the background and the surroundings. Crack pixels have lower intensity than other pixels~\cite{zou2012cracktree}. Consequently, intensity thresholds based methods~\cite{tang201, yu2019dark} are proposed for crack detection. For instance, Tang \etal~\cite{tang201} select the optimal threshold by calculating the median and variance of the histogram of crack images. Yu \etal~\cite{yu2019dark} introduce a low-to-high threshold method based on the neighboring effect and incorporating the location and intensity of dark targets. Additionally, since cracks often resemble a continuous group of pixels, exhibiting topological and morphological similarities to edges. Researchers use edge detection algorithms to detect cracks. Many works utilize edge detection operators such as the Canny detector~\cite{canny1986computational} for crack detection.
Several works adopt frequency domain filters, such as Fast Fourier Transform (FFT), Gabor, wavelet transforms, \etc. Furthermore, some machine learning based methods, such as Support Vector Machines (SVM)~\cite{tabatabaei2019automatic}, Random Forests~\cite{Crackforest}, and AdaBoost~\cite{adaboost}, have also been employed for crack detection.

\subsection{Deep-learning based crack detection}
Recently, thanks to the powerful performance of deep neural networks~\cite{UNET, segnet}, deep learning based methods~\cite{barisin2022methods,fang2020novel} become the mainstream approach for crack detection tasks. Mei \etal~\cite{mei2020densely} propose a deep semantic network based on U-Net~\cite{UNET} with a loss function designed to enhance crack connectivity for road surface crack extraction. Zou \etal propose DeepCrack~\cite{deepcrack} based on the SegNet~\cite{segnet} architecture. DeepCrack involves pairing the convolutional features generated in the encoder network with corresponding features generated at the same scale in the decoder network. FPHBN~\cite{FPHBN} applies a novel loss function with a weighted fusion layer to automatically handle the imbalance between foreground crack pixels and background non-crack pixels. CASA-Net~\cite{CASA-Net} and CrackDet \cite{CrackDet} take bounding box as a crack representation. CrackDet model cracks as a series of oriented sub-cracks and proposes an oriented sub-crack detector, which is based on piecewise angle definition. Bang \etal~\cite{bang2019encoder} and Park \etal~\cite{park2019patch} emplopy crack detection on street scenes images, where the scenes are more complex. Besides, Chen \etal~\cite{chen2024mind} proposed a two-phase clustering-inspired representation learning framework called CIRL, which utilizes an ambiguity-aware segmentation loss to force the network to learn the segmentation variance of each pixel, thereby enabling more precise segmentation of the blurry areas at the edges of cracks. In addition, Lu \etal~\cite{lu2025deep} design a two-stage framework combining YOLO-DEW for coarse detection and U-Net-based segmentation for refined geometry extraction and skeletonization of micro cracks on dam surfaces. To further boost the representation of fine-grained crack patterns, Tong \etal~\cite{tong2025modulated} propose MDCGCN, a novel graph-based deformable convolution module that replaces the standard CNN-based offset generation with a graph convolutional mechanism. This enhances the feature adaptability to complex crack geometries on rail surfaces.

Taking into consideration the ability of transformer models to capture long-range dependencies, CrackFormer ~\cite{liu2021crackformer} introduces a transformer network for crack detection. CrackFormer is primarily composed of SegNet~\cite{segnet} and attention modules, which is good at detecting fine-grained cracks. LECSFormer~\cite{chen2022refined} and CrackViT~\cite{quan2023crackvit} also aim to enhance the capability to model long-range dependencies by optimizing network architectures, thereby improving the ability to detect complex and tiny cracks. Kuang \etal~\cite{kuang2024universal} introduce the Visual Crack Prompt mechanism, which refines the focus of the pre-trained model on high-frequency features, significantly enhancing the model's ability to identify and segment specific crack features. Ma \etal~\cite{ma2024transformer} propose an Information Complementary Fusion module that effectively fuses global and local features, leveraging the advantages of both CNN and transformer-based encoding schemes for complementary strengths. Building on this, they introduce a multi-dimensional attention module for optimizing fused features, enhancing the network's ability to capture long-range dependencies through multidimensional attention operations.  Zhao \etal ~\cite{zhao2024crack} divide crack images into smaller patches, which reduces the computational load while allowing the transformer network to focus more on the crack regions. Yu \etal ~\cite{yu2024robust} propose a robust pavement crack segmentation network based on Swin-Transformer. They introduce a feature pyramid pooling module to provide global priors and design a dual-branch decoder to preserve and learn semantic information, enabling network to handle large-scale images and long-span cracks. Hu \etal~\cite{hu2025ccdformer} propose CCDFormer, a dual-backbone Transformer model for enhanced feature extraction in complex crack detection. CCDFormer utilizes parallel CNN and Transformer branches to capture both local and global crack features, aiming to improve crack detection accuracy by effectively fusing local and global information and addressing the limitations of existing methods in challenging real-world scenarios. Furthermore, DCUFormer~\cite{shan2025dcuformer} enhances transformer-based segmentation by integrating dual cross-attention and upsampling attention mechanisms. It preserves both global semantic context and low-level structural details, achieving accurate segmentation of pavement cracks under complex backgrounds.

In addition to CNNs and Transformers, recent efforts explore new lightweight architectures. For instance, Wei \etal~\cite{liu2025scsegamba} propose SCSegamba, a structure-aware vision Mamba model for pixel-level crack segmentation. It introduces a gated bottleneck convolution and a snake attention scan strategy to efficiently model long-range structural dependencies with low computational cost, enabling deployment on resource-constrained devices. Furthermore, Zhang \etal~\cite{zhang2025crack} developed DCCM-Net, a crack segmentation network that uses a difference convolution-based encoder and a hybrid CNN-Mamba multi-scale attention mechanism to address the challenges of accurately detecting thin, long, and irregular cracks.  The network employs an enhanced convolution module  to extract detailed edge information and a mixed convolution and Mamba attention module to capture both spatial and long-range dependencies, improving the precision of crack segmentation.

Though these existing methods have achieved excellent in-dataset crack detection performance, the cross-dataset performance is often unsatisfied due to changes in image appearance, leading to poor transferability. The proposed CrackCue aims to tackle this limitation in unsatisfied generalization ability. We introduce a simple yet novel concept of robust crack cue to effectively guide the crack detection networks to achieve generalizable and robust crack detection across diverse datasets. The proposed CrackCue is a plug-and-play method that can be incorporated into most existing crack detection methods.
\subsection{Robust curvilinear object segmentation}
As far as we know, there are no other methods specifically designed to enhance the robustness of crack detection. However, due to the slender curvilinear structure of cracks, methods aimed at improving the robustness of curvilinear object segmentation are also effective for cracks. LIOT~\cite{liot} proposes the local intensity order transformation aimed at enhancing the generalization ability of curvilinear object segmentation. This transformation converts grayscale images into four-channel images with contrast-invariant intensity orders based on each pixel and its surrounding pixels along four directions (horizontal and vertical). This results in a representation that preserves the inherent characteristic of curvilinear structures while being robust to changes in contrast. However, in crack detection task, the presence of shadows may affect the relative intensity relationships between pixels in the crack area and the background area. Relying solely on the relative grayscale intensity relationships, LIOT may lead to false detections. Additionally, LIOT lacks robustness to perturbations that can modify the relative brightness of local areas, such as salt-and-pepper noise.
	
\section{Method}
\begin{figure*}[t]
  \centering
  \includegraphics[width=\linewidth]{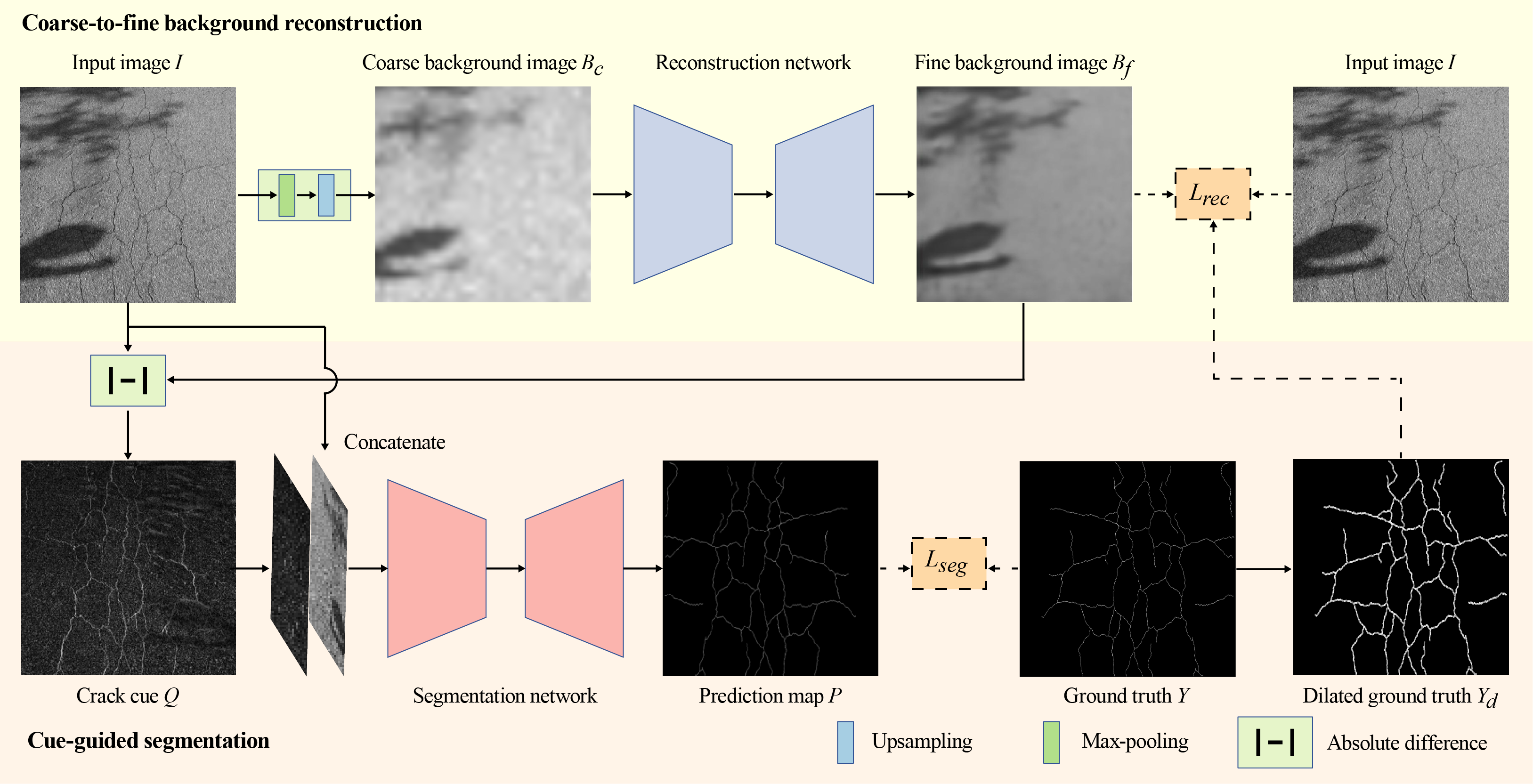}
  \caption{Pipeline of the proposed method CrackCue. We first adopt a simple coarse-to-fine background reconstruction module to obtain the fine background, using reconstruction loss on the background area (inverse of dilated ground-truth crack). The absolute difference between the original image and reconstructed fine background forms the crack cue map. We then feed the concatenation of crack cue and original image into a segmentation network trained with a segmentation loss.}
  \label{fig:pipeline}
\end{figure*}
\subsection{Overview}
The overall pipeline of the proposed CrackCue is illustrated in Figure~\ref{fig:pipeline}. First, the coarse-to-fine background reconstruction module reconstructs a crack-free background image from the crack image. Subsequently, the absolute difference between the background image and the original image is calculated to generate a crack cue. This cue is then fed into the cue-guided segmentation module for robust crack detection. We detail these two modules in the following. For clarity, a simplified version of the pipeline is also illustrated in Figure~\ref{fig:flowchart}.

\subsection{Coarse-to-fine background reconstruction}
Cracks typically exhibit a thin structure and are often darker compared to the surrounding background areas. Therefore, the input image $I$ is passed through a max-pooling layer with a kernel size of 8. Since the intensity of crack pixels is relatively low, most of them are eliminated. Then, an upsampling operation with a scale of 8 is applied to restore the image to its original resolution.
This yields a coarse background image $B_c$ without cracks. However, the max-pooling operation may lead to significant texture detail loss, resulting in quite blurry coarse background image $B_c$. Directly generating crack cue by comparing the difference between $B_c$ and the input image $I$ results in numerous false positives in the background areas. To cope with this issue, we employ a reconstruction network to refine $B_c$, aiming to generate a more detailed background image.

The reconstruction network is formulated as an encoder-decoder structure, capable of transforming the coarse background image $B_c$ into a fine background image $B_f$. To prevent crack features from leaking into the decoder network through skip connections, which could result in incomplete crack elimination, the skip connections are removed. Both the encoder and decoder consist of 10 convolution layers.
We select the background region of the input image $I$ as the reconstruction target to recover the lost texture details in the background region of $B_c$. To prevent the network from reconstructing crack pixels, we use the ground truth label $Y$ to mask the crack regions in the input image $I$ when calculating the reconstruction loss. Since the annotations in $Y$ are only a single pixel-width, insufficient to cover the actual crack width in $I$, we expand $Y$ into a dilated version $Y_d$ with width $d$, to adequately cover the cracks. The final reconstruction loss $L_{rec}$ is calculated by:
\begin{equation}
L_{rec} = \frac{1}{N} \sum \left(\left(1-Y_d\right) \odot \left(I - B_f\right)^2\right),
\label{final recon loss}
\end{equation}
where $\odot$ denotes the element-wise multiplication of two matrices, and $N = H\times W$ with $H$ and $W$ representing corresponding height and width of $I$, respectively. In $L_{rec}$, there are no specific constraints applied to the crack areas and the input of the reconstruction network $B_c$ does not contain crack pixels. The pixels in $B_f$ corresponding to the crack areas tend to remain homogenous with the surrounding background. This results in a more pronounced difference in the crack areas between $B_f$ and $I$.

Through the aforementioned process, our reconstruction module achieves a coarse-to-fine reconstruction of the crack-free background images $B_f$. $B_f$ is then fed into the cue-guided segmentation module to obtain crack cue for subsequent crack detection task.

\subsection{Cue-guided segmentation}
The crack cue $Q$ used for guiding the crack segmentation is derived by calculating the absolute difference between the input image $I$ and the crack-free background image $B_f$, expressed as 
\begin{equation}
Q = \frac{1}{3} \sum_{i=1}^{3} \left| I^i - B_{f}^i \right|,
\label{Crack cue}
\end{equation}
where $i\in \{1,2,3\}$ represents the image channel dimension index. Subsequently, $Q$ is concatenated with $I$ along the channel dimension to form the input $I_{seg}$ for the segmentation network.
Given that $B_f$ effectively reconstructs the background area of the input image $I$ while excluding crack pixels, there is a notable discrepancy between them in the crack regions. Consequently, the  crack cue $Q$ exhibits high intensity in the crack regions, providing rich prior information for the crack detection task. Changes in lighting conditions or background textures similarly affect both the input image $I$ and the reconstructed background image $B_f$. This similarity allows the generated crack cue $Q$, which is the absolute difference between $I$ and $B_f$, to effectively mitigate these disturbances. As a result, this approach equips the proposed CrackCue with a high level of robustness against varying environmental conditions.

\begin{figure}[t]
  \centering
  \includegraphics{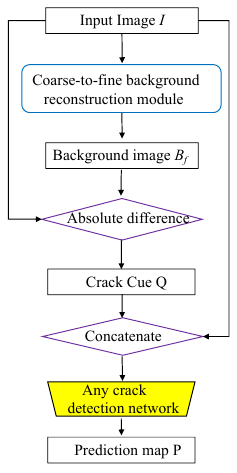}
  \caption{The flowchart of the proposed CrackCue framework. The crack detection network in the yellow box can be replaced with any existing crack segmentation network.}
  \label{fig:flowchart}
\end{figure}

As a plug-and-play module, the proposed cue-guided segmentation can be flexibly integrated with most crack detection networks by replacing the segmentation network (i.e., swapping the yellow box in the flowchart shown in Figure~\ref{fig:flowchart} with the network) in the process. We apply Binary CrossEntropy Loss as $L_{seg}$ to supervise the segmentation network, denoted as 
\begin{equation}
L_{seg} = L_{bce}\left(Y, P\right),
\label{BCE loss}
\end{equation}
where $Y$ represents the ground truth, and $P$ denotes the prediction map of the segmentation network. For networks requiring additional loss function to constraint (such as those outputting predictions at multiple scales), we follow their specific configurations to adjust $L_{seg}$.

During the training stage, we jointly train both reconstruction and segmentation networks. The final total loss $L$ is defined as:
\begin{equation}
L = L_{seg} + \lambda \times L_{rec}.
\label{total loss}
\end{equation} 
where $\lambda$ balances the two loss function terms. During the testing stage, the test image is fed into the coarse-to-fine background reconstruction module. The crack cue is obtained by calculating the absolute difference between the reconstructed image and the test image. Then, the test image is concatenated with the crack cue along the channel dimension and input into the cue-guided segmentation module. The output of the segmentation model is normalized to a range of 0 to 1 through a sigmoid function, yielding the final prediction probability map.

\section{Experiments}
\subsection{Datasets and evalution metrics}
In this study, four publicly available crack datasets are used. Following the setting in DeepCrack~\cite{deepcrack}, the pavement crack dataset CrackTree260\footnote{https://1drv.ms/f/s!AittnGm6vRKLyiQUk3ViLu8L9Wzb}~\cite{zou2012cracktree} is utilized as the training set, while the other three datasets\footnote{https://1drv.ms/f/s!AittnGm6vRKLtylBkxVXw5arGn6R} are employed to test the generalization performance of the network. Each dataset provides detailed pixel-level annotations.

\medskip
\noindent\textbf{CrackTree260}~\cite{zou2012cracktree} comprises 260 road pavement images with a resolution of $800\times600$, captured by an area array camera under visible light illumination.  For training purposes, we resize these images to a resolution of $512\times512$.

\medskip
\noindent\textbf{CrackLS315}~\cite{deepcrack} includes 315 asphalt road surface images captured by a line scan camera under laser illumination, with an image resolution of $512\times512$. 

\medskip
\noindent\textbf{CRKWH100}~\cite{deepcrack} contains 100 road pavement images taken with a line scan camera under visible light illumination at a ground sampling distance of 1 millimeter. Each image is also of size $512\times512$.

\medskip
\noindent\textbf{Stone331}~\cite{deepcrack} contains 331 images of stone surfaces, captured by an area array camera under visible light illumination. The original size of the images is $1024\times1024$. They are resized to $512\times512$ for testing purposes. Stone331 also provides a corresponding mask for each image, ensuring that performance evaluation is focused solely on the stone surface.

\medskip
\noindent\textbf{Evaluation metrics:}
We follow the setup of DeepCrack~\cite{deepcrack} for quantitative evaluation. For each image, we calculate precision and recall by comparing ground truth with segmentation results, and then compute the F1 score. The F1 score, the harmonic mean of precision and recall, is given by $2 \times \frac{\text{precision} \times \text{recall}}{\text{precision} + \text{recall}}$.
Specifically, three different F-measure-based metrics are used in the evaluation: Optimal Dataset Scale (ODS), which corresponds to the best F-measure on the dataset at a fixed threshold; Optimal Image Scale (OIS), which aggregates the best F-measure at the best threshold for each image on the dataset; and Average Precision (AP), which is the area under the precision-recall curve. Considering that cracks in images have a certain width, we set a tolerance of 3 pixels.

\subsection{Implementation details}
Our method is implemented using the open-source framework PyTorch\footnote{https://pytorch.org/}. We jointly train the reconstruction and segmentation networks with a batch size of 4, over a total of 700 epochs. The Adam optimizer is employed to optimize the network, with the learning rate set to 1e-4. For the reconstruction module, the kernel size of the max-pooling layers is set to 8, and the width $d$ of dilated mask $Y_d$ in the computation of $L_{rec}$ is set to 4. The balancing parameter $\lambda$ for the two loss terms in Eq.~\eqref{total loss} is set to 1. We reproduce all baseline methods using the same settings. All the experiments in this paper are conducted on two GeForce RTX 3090 GPUs.

\begin{table*}[t]
\centering
\caption{Cross-dataset evaluation of CrackCue, LIOT~\cite{liot} and some popular baseline methods on crack detection. The best results within each pair of comparative experiments are bolded. It is noteworthy that the original CrackFormer is evaluated under in-dataset setting. The results in this table are reproduced using the official code\protect\footnotemark.}
 \resizebox{1\textwidth} {!}{
\begin{tabular}{|l|c|c|c|c|c|c|c|c|c|c|c|c|}
\hline
\rule{0pt}{10pt} \multirow{2}{*}{Method} & \multicolumn{3}{c|}{CrackLS315~\cite{deepcrack}} & \multicolumn{3}{c|}{CRKWH100~\cite{deepcrack}} & \multicolumn{3}{c|}{Stone331~\cite{deepcrack}} & \multicolumn{3}{c|}{Average}\\ \cline{2-13} 
\rule{0pt}{10pt} & ODS & OIS & AP & ODS & OIS & AP& ODS & OIS & AP & ODS & OIS & AP\\ \hline
\rule{0pt}{10pt} U-Net~\cite{UNET} & 0.790 & 0.793 & 0.811 & 0.937 & 0.949 & 0.942  &  0.884 & 0.914 & 0.912 &0.870 &0.885 &0.888\\ 
\rule{0pt}{10pt} +LIOT~\cite{liot} & 0.836 & 0.841 & 0.847 & 0.941 & 0.958 & 0.952  &  0.910 & 0.934 & 0.929 &0.896{\footnotesize (+2.6\%)} &0.911{\footnotesize (+2.6\%)} &0.909{\footnotesize (+2.1\%)}\\ 
\rowcolor{gray!20}
\rule{0pt}{10pt} +CrackCue & \textbf{0.847} & \textbf{0.851} & \textbf{0.860}  & \textbf{0.949}  & \textbf{0.968}  & \textbf{0.961}  & \textbf{0.920}  & \textbf{0.947}  & \textbf{0.942} &\textbf{0.905{\footnotesize (+3.5\%)}} &\textbf{0.922{\footnotesize (+3.7\%)}} &\textbf{0.921{\footnotesize (+3.3\%)}}\\ \hline

\rule{0pt}{10pt} CrackFormer~\cite{liu2021crackformer} & 0.811 & 0.845 & 0.828 & 0.934 & 0.960 & 0.951  & 0.752 & 0.820 & 0.757 &0.832 &0.875 &0.845\\ 
\rule{0pt}{10pt} +LIOT~\cite{liot} & 0.877 & 0.909 & 0.896 & 0.944 & 0.961 & 0.954 &  0.821 & 0.865 & 0.831 &0.881{\footnotesize (+4.9\%)} &0.912{\footnotesize (+3.7\%)} &0.894{\footnotesize (+4.9\%)}\\ 
\rowcolor{gray!20}
\rule{0pt}{10pt} +CrackCue & \textbf{0.887}  & \textbf{0.914}  & \textbf{0.899}  & \textbf{0.945}  & \textbf{0.967}  & \textbf{0.958}   & \textbf{0.867}  & \textbf{0.919}  & \textbf{0.903} &\textbf{0.900{\footnotesize (+6.8\%)}} &\textbf{0.933{\footnotesize (+5.8\%)}} &\textbf{0.920{\footnotesize (+7.5\%)}} \\ \hline

\rule{0pt}{10pt} DeepCrack~\cite{deepcrack} & 0.848 & 0.859 & 0.843 & 0.949 & 0.963 & 0.950  & 0.905 & 0.931 & 0.925 &0.901 &0.918 &0.906\\ 
\rule{0pt}{10pt} +LIOT~\cite{liot} & 0.862 & 0.871 & 0.868 & 0.949 & 0.966 & 0.959  &  0.922 & 0.948 & 0.940 &0.911{\footnotesize (+1.0\%)} &0.928{\footnotesize (+1.0\%)} &0.922{\footnotesize (+1.6\%)}\\ 
\rowcolor{gray!20}
\rule{0pt}{10pt} +CrackCue & \textbf{0.890}  & \textbf{0.897}  & \textbf{0.886} & \textbf{0.954} & \textbf{0.970}  & \textbf{0.962}   & \textbf{0.928}  & \textbf{0.959}  & \textbf{0.951}  & \textbf{0.924{\footnotesize (+2.3\%)}} & \textbf{0.942{\footnotesize (+2.4\%)}} & \textbf{0.933{\footnotesize (+2.7\%)}}\\ \hline

\end{tabular}
}

\label{tab1}
\end{table*}
\footnotetext{https://github.com/LouisNUST/CrackFormer-II}

\begin{figure*}[t]
  \centering
  \includegraphics[width=\linewidth]{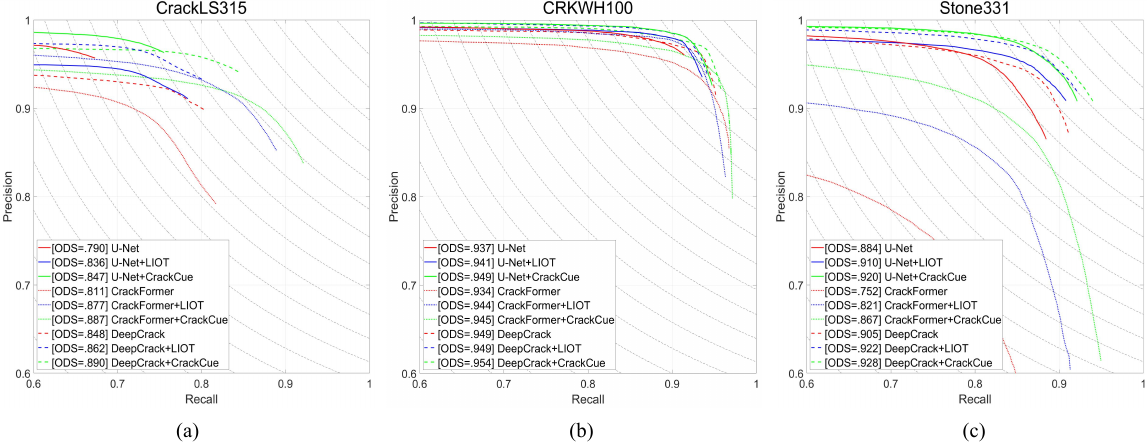}
  \caption{The Precision-Recall curves of the proposed CrackCue and LIOT~\cite{liot} built upon some popular baseline methods on the three test datasets.}
  \label{fig:PR}
\end{figure*}

\begin{table*}
\centering
\caption{Quantitative cross-dataset evaluation of the proposed CrackCue, LIOT~\cite{liot} and some popular baseline methods on crack images with defocus blur and contrast perturbation~\cite{hendrycks2019benchmarking}. The proposed CrackCue is more robust to these perturbations.}
\setlength{\tabcolsep}{1mm}
 \resizebox{1.0\linewidth}{!}{
\begin{tabular}{|c|l  |c|c|c  |c|c|c  |c|c|c|  c|c|c|}
\hline
\rule{0pt}{10pt} \multirow{2}{*}{ Perturbation } & \multirow{2}{*}{Method} & \multicolumn{3}{c|}{CrackLS315~\cite{deepcrack}} & \multicolumn{3}{c|}{CRKWH100~\cite{deepcrack}} & \multicolumn{3}{c|}{Stone331~\cite{deepcrack}} & \multicolumn{3}{c|}{Average}\\ \cline{3-14} 
\rule{0pt}{10pt} & & ODS & OIS & AP & ODS & OIS & AP & ODS & OIS & AP & ODS & OIS & AP \\ \hline

\rule{0pt}{10pt}\multirow{4}{*}{Defocus} &U-Net~\cite{UNET}  &0.531 &0.534 &0.576 &0.834 &0.839  &0.851  &0.852  &0.886  &0.881 &0.739 &0.753 &0.769 \\
\rule{0pt}{10pt} &+LIOT~\cite{liot}  &0.503 &0.573 &0.453 &0.806 &0.880  &0.835 &0.861  &0.913 &0.902 &0.723{\footnotesize (-1.6\%)} &0.789{\footnotesize (+3.6\%)} &0.730{\footnotesize (-3.9\%)}\\

\rule{0pt}{10pt} &\cellcolor{gray!20}+CrackCue  &\cellcolor{gray!20}\textbf{0.614}   &\cellcolor{gray!20}\textbf{0.638}   &\cellcolor{gray!20}\textbf{0.627}  &\cellcolor{gray!20}\textbf{0.881}   &\cellcolor{gray!20}\textbf{0.913}   &\cellcolor{gray!20}\textbf{0.893}   &\cellcolor{gray!20}\textbf{0.889}  &\cellcolor{gray!20}\textbf{0.931}   &\cellcolor{gray!20}\textbf{0.930} &\cellcolor{gray!20}\textbf{0.795{\footnotesize (+5.6\%)}}  &\cellcolor{gray!20}\textbf{0.827{\footnotesize (+7.4\%)}} &\cellcolor{gray!20}\textbf{0.817{\footnotesize (+4.8\%)}}\\ \cline{2-14}

\rule{0pt}{10pt} \multirow{4}{*}{blur} &CrackFormer~\cite{liu2021crackformer} &0.584 &0.611  &0.538  &0.820  &0.860  &0.824  &0.686 &0.762  &0.662 &0.697 &0.744 &0.675\\
\rule{0pt}{10pt} &+LIOT~\cite{liot}  &0.561 &0.592 &0.501 &0.828 &0.854  &0.825  &0.402  &0.443 &0.293 &0.597{\footnotesize (-10.0\%)} &0.630{\footnotesize (-11.4\%)} &0.540{\footnotesize (-13.5\%)}\\

\rule{0pt}{10pt} &\cellcolor{gray!20}+CrackCue  &\cellcolor{gray!20}\textbf{0.607}   &\cellcolor{gray!20}\textbf{0.645}  &\cellcolor{gray!20}\textbf{0.539}   &\cellcolor{gray!20}\textbf{0.863}   &\cellcolor{gray!20}\textbf{0.902}   &\cellcolor{gray!20}\textbf{0.875}   &\cellcolor{gray!20}\textbf{0.784} &\cellcolor{gray!20}\textbf{0.845}   &\cellcolor{gray!20}\textbf{0.805} &\cellcolor{gray!20}\textbf{0.751{\footnotesize (+5.4\%)}} &\cellcolor{gray!20}\textbf{0.797{\footnotesize (+5.3\%)}} &\cellcolor{gray!20}\textbf{0.740{\footnotesize (+6.5\%)}}\\ \cline{2-14}

 \rule{0pt}{10pt}  &DeepCrack~\cite{deepcrack} &0.533  &0.537  &0.602  &0.857  &0.868  &0.871  &0.873  &0.903  &0.894 &0.754 &0.769 &0.789\\
 \rule{0pt}{10pt} &+LIOT~\cite{liot}  &0.556 &0.596 &0.538 &0.860 &0.905  &0.890  &0.861  &0.910  &0.897 &0.759{\footnotesize (+0.5\%)} &0.804{\footnotesize (+3.5\%)} &0.775(-1.4\%)\\

\rule{0pt}{10pt} &\cellcolor{gray!20}+CrackCue  &\cellcolor{gray!20}\textbf{0.652}   &\cellcolor{gray!20}\textbf{0.678}   &\cellcolor{gray!20}\textbf{0.676}   &\cellcolor{gray!20}\textbf{0.887}   &\cellcolor{gray!20}\textbf{0.935}   &\cellcolor{gray!20}\textbf{0.934}   &\cellcolor{gray!20}\textbf{0.891}   &\cellcolor{gray!20}\textbf{0.936}   &\cellcolor{gray!20}\textbf{0.929} &\cellcolor{gray!20}\textbf{0.810{\footnotesize (+5.6\%)}} &\cellcolor{gray!20}\textbf{0.850{\footnotesize (+8.1\%)}} &\cellcolor{gray!20}\textbf{0.846{\footnotesize (+5.7\%)}}\\ \cline{2-14}

\hline

\rule{0pt}{10pt} \multirow{9}{*}{Contrast} &U-Net~\cite{UNET} &0.523  &0.524  &0.636  &0.724  &0.725  &0.777  &0.757  &0.758  &0.800 &0.668 &0.669 &0.738\\
 \rule{0pt}{10pt} &+LIOT~\cite{liot}  &0.682 &0.709 &0.695 &0.904 &0.923  &0.920  &0.856  &0.891  &0.885 &0.814{\footnotesize (+14.6\%)} &0.841{\footnotesize (+17.2\%)} &0.833{\footnotesize (+9.5\%)}\\

\rule{0pt}{10pt} &\cellcolor{gray!20}+CrackCue  &\cellcolor{gray!20}\textbf{0.734}   &\cellcolor{gray!20}\textbf{0.763}   &\cellcolor{gray!20}\textbf{0.755}   &\cellcolor{gray!20}\textbf{0.908}   &\cellcolor{gray!20}\textbf{0.940}   &\cellcolor{gray!20}\textbf{0.941}   &\cellcolor{gray!20}\textbf{0.885}   &\cellcolor{gray!20}\textbf{0.915}   &\cellcolor{gray!20}\textbf{0.916} &\cellcolor{gray!20}\textbf{0.842{\footnotesize (+17.4\%)}} &\cellcolor{gray!20}\textbf{0.873{\footnotesize (+20.4\%)}} &\cellcolor{gray!20}\textbf{0.871{\footnotesize (+13.3\%)}}\\ \cline{2-14}

\rule{0pt}{10pt} &CrackFormer~\cite{liu2021crackformer} &0.665 &0.705 &0.681  &0.856  &0.902  &0.895  &0.711  &0.725  &0.751 &0.744 &0.777 &0.776\\
 \rule{0pt}{10pt} &+LIOT~\cite{liot}  &0.706 &0.743 &0.724 &0.884 &0.912  &0.908  &0.714  &0.745  &0.749 &0.768{\footnotesize (+2.4\%)} &0.800{\footnotesize (+2.3\%)} &0.794{\footnotesize (+1.8\%)}\\

\rule{0pt}{10pt} &\cellcolor{gray!20}+CrackCue  &\cellcolor{gray!20}\textbf{0.773}   &\cellcolor{gray!20}\textbf{0.795}   &\cellcolor{gray!20}\textbf{0.788}   &\cellcolor{gray!20}\textbf{0.921}   &\cellcolor{gray!20}\textbf{0.944}  &\cellcolor{gray!20}\textbf{0.940}  &\cellcolor{gray!20}\textbf{0.744}  &\cellcolor{gray!20}\textbf{0.801}   &\cellcolor{gray!20}\textbf{0.761} &\cellcolor{gray!20}\textbf{0.813{\footnotesize (+6.9\%)}} &\cellcolor{gray!20}\textbf{0.847{\footnotesize (+7.0\%)}} &\cellcolor{gray!20}\textbf{0.830{\footnotesize (+5.4\%)}}\\ \cline{2-14}

\rule{0pt}{10pt} &DeepCrack~\cite{deepcrack}  &0.510  &0.511  &0.616  &0.541  &0.541  &0.680  &0.570  &0.571  &0.697 &0.540 &0.541 &0.664\\
 \rule{0pt}{10pt} &+LIOT~\cite{liot}  &0.706 &0.742 &0.713 &0.864 &0.872  &0.866  &0.754  &0.788  &0.792 &0.775{\footnotesize (+23.5\%)} &0.801{\footnotesize (+26.0\%)} &0.790{\footnotesize (+12.6\%)}\\

\rule{0pt}{10pt} &\cellcolor{gray!20}+CrackCue  &\cellcolor{gray!20}\textbf{0.798}  &\cellcolor{gray!20}\textbf{0.809}  &\cellcolor{gray!20}\textbf{0.788}  &\cellcolor{gray!20}\textbf{0.901}   &\cellcolor{gray!20}\textbf{0.928}   &\cellcolor{gray!20}\textbf{0.923}   &\cellcolor{gray!20}\textbf{0.893}   &\cellcolor{gray!20}\textbf{0.902}   &\cellcolor{gray!20}\textbf{0.903} &\cellcolor{gray!20}\textbf{0.864{\footnotesize (+32.4\%)}} &\cellcolor{gray!20}\textbf{0.880{\footnotesize (+33.9\%)}} &\cellcolor{gray!20}\textbf{0.871{\footnotesize (+20.7\%)}}\\ 

\hline
\end{tabular}
 }
\label{tab6}
\end{table*}

\begin{figure*}[t]
  \centering
  \includegraphics[width=\linewidth]{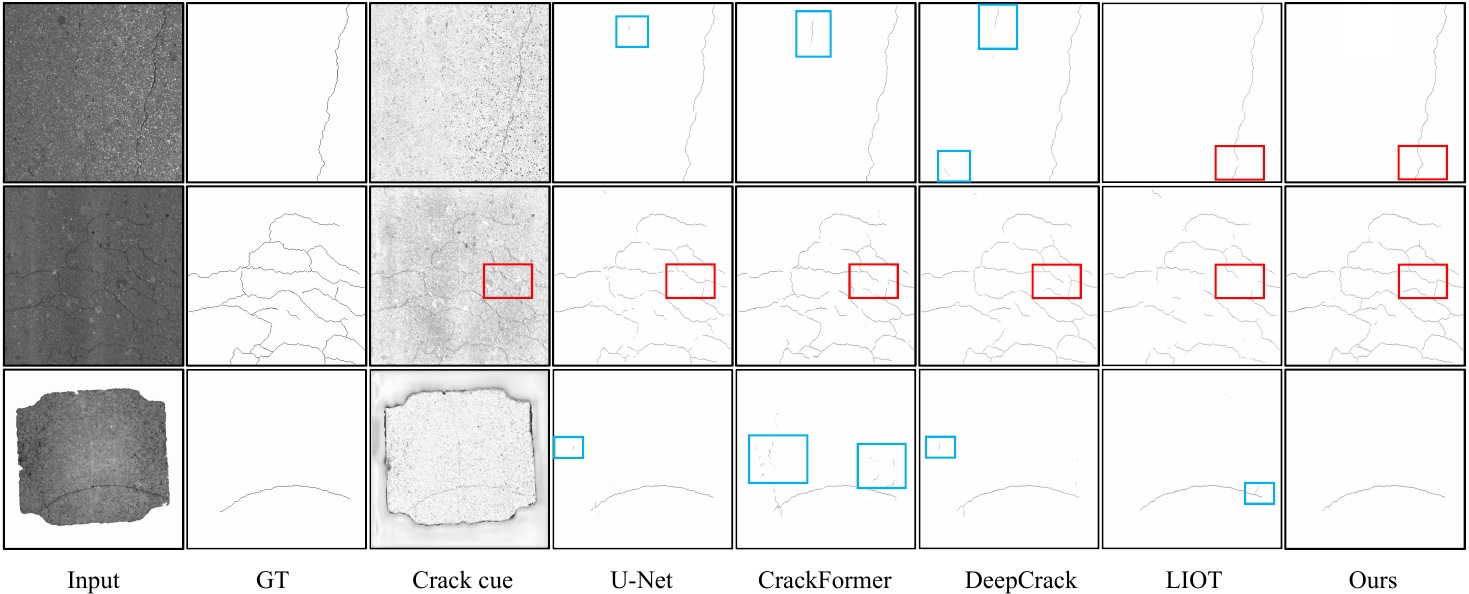}
  \caption{Qualitative comparison of crack detection results between some popular baseline methods and the proposed CrackCue built upon DeepCrack~\cite{deepcrack}, as well as LIOT~\cite{liot}, on three sample images (from top to bottom) selected from the CrackLS315, CRKWH100, and Stone331 dataset. For better visibility, we perform a grayscale inversion on the GTs, crack cues, and the predicted results.
  Blue boxes indicate false positives, while red boxes represent false negatives. }
  \label{fig:vis}
\end{figure*}

\subsection{Cross-crack-dataset evaluation}
To validate the effectiveness of the proposed plug-and-play CrackCue, we select three networks that have demonstrated significant performance and are widely compared in crack detection tasks: U-Net \cite{UNET}, CrackFormer \cite{liu2021crackformer}, and DeepCrack \cite{deepcrack}. The first two are representative of CNN-based crack detection networks, while CrackFormer represents Transformer-based crack detection networks.
We establish baselines by directly segmenting input crack images using the three networks. Subsequently, we compare these baselines with the performance of the same networks when enhanced by crack cues generated by CrackCue. Additionally, since cracks are a typical linear structure, we compared the proposed CrackCue algorithm with LIOT~\cite{liot}, an algorithm designed specifically for robust curve object segmentation, to demonstrate the superiority of CrackCue in crack detection tasks.
All experiments are trained on the CrackTree dataset and evaluated for cross-dataset performance on the other three datasets. It is worth noting that the original CrackFormer is a state-of-the-art (SOTA) method evaluated under in-dataset setting, which differs from our cross-dataset setup.

Figure~\ref{fig:vis} presents a series of visualization results for comparison. The distinct response of crack cues in crack areas clearly demonstrates how they provide essential prior information, enhancing crack segmentation performance. Observing the final segmentation results, the introduction of CrackCue improves segmentation by reducing false positives (blue boxes in Figure~\ref{fig:vis}) and missed detections, leading to cracks with better connectivity (as shown in the red box in Figure~\ref{fig:vis}). In general, the proposed CrackCue is effective in improving the detection accuracy. The quantitative results are presented in Table~\ref{tab1}.
The performance of all three baseline networks shows significant improvement when integrated with LIOT or CrackCue. However, comparatively, the enhancement provided by the proposed CrackCue is more pronounced. Compared to the three baseline networks, the average improvements in ODS across the three datasets reach 3.5\%, 6.8\%, and 2.3\%, respectively. The PR curves in Figure~\ref{fig:PR} demonstrate that, for each network, the performance with CrackCue outperforms LIOT and the baseline itself on all three datasets (\textit{i.e.}, curves of the same shape, with the green curve consistently closer to the upper right corner). This validates the effectiveness of CrackCue and its ability to generalize across images with different appearances. Further detailed analysis of the experimental results on the three datasets is provided below.

\medskip
\noindent\textbf{1) CrackLS315}~\cite{deepcrack}: Crack detection in the CrackLS315 dataset is challenging due to its extremely low image contrast from laser illumination. As illustrated in Figure~\ref{fig:PR}(a), the performance of the three baseline networks on the CrackLS315 dataset is suboptimal. However, after employing CrackCue to guide the segmentation, the PR curves of all three networks significantly shifted upwards, resulting in a noticeable enhancement in performance. The quantitative results in Table~\ref{tab1} are more illustrative. After integrating CrackCue, the ODS scores of the three networks increased by 5.7\%, 7.6\%, and 4.2\%, respectively. This improvement arises due to the limited generalization capability of segmentation networks. They struggle particularly with significant appearance deviations between training and testing images, especially under varying conditions. Yet, the crack cues generated by the proposed CrackCue are less sensitive to changes in image contrast and provide robust prior information for the segmentation network.

\medskip
 \noindent\textbf{2) CRKWH100}~\cite{deepcrack}: The PR curves in Figure~\ref{fig:PR}(b) indicate high baseline results, likely thanks to favorable lighting and simpler structures in the CRKWH100 dataset. However, upon incorporating CrackCue for segmentation guidance, the PR curves all advance towards the upper right corner, which is particularly evident in the cases of U-Net~\cite{UNET} and CrackFormer~\cite{liu2023crackformer}. This enhancement is quantitatively confirmed in Table~\ref{tab1}, where the ODS scores for U-Net~\cite{UNET}, DeepCrack~\cite{deepcrack}, and CrackFormer~\cite{liu2021crackformer} increase by 1.2\%, 1.1\%, and 0.5\%, respectively. These gains are appreciable, considering the already high baseline ODS scores above 93\%. This also demonstrates the effectiveness of the crack cue from our CrackCue in providing valuable prior information.

\medskip
\noindent\textbf{3) Stone331}~\cite{deepcrack}: Stone331 originates from stone surfaces, presenting a significant domain gap due to smoother textures compared to road surface cracks. Figure~\ref{fig:PR}(c) and Table~\ref{tab1} show that CrackFormer~\cite{liu2023crackformer}, despite its effectiveness in the first two datasets, experiences a significant drop in performance on stone surfaces. Yet, guiding CrackFormer with CrackCue substantially improves its PR curve and boosts the ODS by 11.5\%, effectively reducing the impact of domain variation. Despite U-Net~\cite{UNET} and DeepCrack~\cite{deepcrack} showing stronger cross-domain capabilities in these experiments, incorporating CrackCue resulted in ODS improvements of 3.6\% and 2.3\% respectively, highlighting the consistent effectiveness of the proposed CrackCue.

\subsection{Cross-dataset evaluation on perturbed crack images}
It is well-known that the performance of deep learning-based models are susceptible to image perturbations. In this section, we evaluate the robustness of the proposed CrackCue against various perturbations. For this purpose, we adopt the approach described in \cite{hendrycks2019benchmarking} to add two common types of disturbances to crack images, namely low contrast and defocus blur. Under such disturbances, we conduct experiments to compare the individual performance of three baseline networks separately and their performance when combined with LIOT~\cite{liot} and CrackCue. The quantitative results in Table~\ref{tab6} indicate that, in the majority of cases, CrackCue can alleviate the impact of perturbations on crack detection networks to varying degrees. For crack segmentation under low contrast disturbances, networks integrated with CrackCue demonstrate a noticeable improvement in ODS. Compared to the three baselines, the average enhancements reach substantial levels of 17.4\%, 6.9\%, and 32.0\%, respectively. Furthermore, compared to LIOT, CrackCue also achieves much better performance gains. Additionally, while defocus blur can cause changes in the local pixel brightness order, LIOT performs poorly under this disturbance, whereas CrackCue maintains high performance. The extensive experimental results in Table~\ref{tab6} effectively illustrate the robustness of the proposed CrackCue against interference, underlining its practical value for crack detection tasks in real-world scenarios.

\subsection{Ablation study}

\begin{table}[t]
\caption{Ablation study on the effect of using variants of crack cues for the proposed CrackCue built upon U-Net~\cite{UNET}.}
\centering
 \resizebox{1.0\linewidth}{!}{
\begin{tabular}{|l|c|c|c|}
\hline
Method  & ODS & OIS & AP \\ \hline
U-Net~\cite{UNET}  & 0.790 & 0.793 & 0.811 \\ \hline
+Cue with median filter &0.816 {\footnotesize (+2.6\%)} &0.820  {\footnotesize (+2.7\%)}&0.828 {\footnotesize (+1.7\%)}\\ \hline
+Cue with black-hat operator&0.799  {\footnotesize (+0.9\%)} &0.802 {\footnotesize (+0.9\%)} &0.813 {\footnotesize (+0.2\%)}\\ \hline
+Cue with Canny detector &0.779 {\footnotesize (-1.1\%)} &0.782  {\footnotesize (-1.1\%)}&0.799 {\footnotesize (-1.2\%)}\\ \hline
\rowcolor{gray!20}
+Coarse crack cue & 0.833 {\footnotesize (+4.3\%)} & 0.836 {\footnotesize (+4.3\%)} & 0.846 {\footnotesize (+3.5\%)} \\ \hline
\rowcolor{gray!20}
+Fine crack cue  & \textbf{0.847} {\footnotesize (+5.7\%)} & \textbf{0.851} {\footnotesize (+5.8\%)}& \textbf{0.860} {\footnotesize (+4.9\%)} \\ \hline
\end{tabular}
}
\label{tab3}
\end{table}

\begin{table}[t]
\caption{Ablation study on the hyper-parameter $d$ in Eq.~\eqref{final recon loss} for dilated GT $Y_d$.}
\centering
\begin{tabular}{|c|c|c|c|}
\hline
Width $d$ & ODS & OIS & AP\\ 
\hline
2 & 0.837 & 0.841 & 0.851\\ 
\hline
3 & 0.846 & 0.850 & 0.859\\ 
\hline
\rowcolor{gray!20}
4 & \textbf{0.847} & \textbf{0.851} & \textbf{0.860}\\ 
\hline
5 & 0.841 & 0.844 & 0.854\\ 
\hline
\end{tabular}
\label{tab5}
\end{table}

\begin{table}[t]
\centering
\caption{Ablation study on the balancing parameter $\lambda$ in Eq.~\eqref{total loss}.}

\begin{tabular}{|l|c|c|c|}
\hline
 Method & ODS & OIS & AP \\ \hline
 U-Net + CrackCue ($\lambda$ = 0.25) & 0.841 & 0.847 & 0.851 \\ \hline
 \rowcolor{gray!20}
 U-Net + CrackCue ($\lambda$ = 1) & \textbf{0.847} & \textbf{0.851} & \textbf{0.860} \\ \hline
 U-Net + CrackCue ($\lambda$ = 4) & 0.839 & 0.848 & 0.849 \\ \hline
\end{tabular}

\label{tab:lambda}
\end{table}

\begin{table}[t]
\centering
\caption{Ablation study on the kernel size of max-pooling operator.}
\begin{tabular}{|l|c|c|c|}
\hline
 kernel size & ODS & OIS & AP \\ \hline
 4 & 0.825 & 0.839 & 0.844 \\ \hline
 \rowcolor{gray!20}
 8 & \textbf{0.847} & \textbf{0.851} & \textbf{0.860} \\ \hline
 12 & 0.819 & 0.836 & 0.842 \\ \hline
\end{tabular}
\label{tab:kernel}
\end{table}

\begin{table}[t]
\caption{Runtime comparison between the proposed CrackCue and baseline networks.}
\centering
\resizebox{1.0\linewidth}{!}{
\begin{tabularx}{\linewidth}{|>{\raggedright\arraybackslash}X|>{\centering\arraybackslash}X|>{\centering\arraybackslash}X|}
\hline
\rule{0pt}{6pt} Method  & ODS & FPS \\ 
\hline
U-Net~\cite{UNET}  & 0.790 & \textbf{25.3}  \\ 
\rowcolor{gray!20}
+Coarse cue &0.833{\footnotesize (+4.3\%)} & 25.3 \\ 
\rowcolor{gray!20}
+CrackCue & \textbf{0.847} {\footnotesize (+5.7\%)} & 11.2    \\ 
\hline
CrackFormer~\cite{liu2021crackformer} &0.811  & \textbf{7.9}\\ 
\rowcolor{gray!20}
+Coarse cue &0.850{\footnotesize (+3.9\%)} & 7.9 \\ 
\rowcolor{gray!20}
+CrackCue &\textbf{0.887} {\footnotesize (+7.6\%)} &7.0 \\ 
\hline
DeepCrack~\cite{deepcrack}  & 0.848 & \textbf{11.0}  \\ 
\rowcolor{gray!20}
+Coarse cue &0.872{\footnotesize (+2.4\%)} & 11.0 \\ 
\rowcolor{gray!20}
+CrackCue &\textbf{0.890} {\footnotesize (+4.2\%)} & 7.7\\ 
\hline
\end{tabularx}
}
\label{runtime}
\end{table}

We perform three types of ablation studies. The first one evaluates the effectiveness of coarse-to-fine background reconstruction by analyzing the impact of both coarse and fine cues generated by the proposed CrackCue on crack detection. The second compares CrackCue with other variants of cue generation. The third examines the effect of the width of $Y_d$ in $L_{rec}$. All experiments are conducted using U-Net~\cite{UNET} as the baseline and are validated on the CrackLS315~\cite{deepcrack} dataset, as it presents the most challenging scenarios.

\medskip
\noindent\textbf{Effect of coarse-to-fine background reconstruction.} Our proposed coarse-to-fine background reconstruction module generates background images of coarse and fine texture, naturally leading to cue of varying degrees. To evaluate their effectiveness, we compare the performance under three conditions: segmentation guided by coarse cue, segmentation guided by fine cue, and direct segmentation of the input image. As shown in Table~\ref{tab3}, segmentation guided by coarse cue significantly enhances performance compared to the baseline. This improvement is thanks to the high response of the coarse cue in crack areas, as crack pixels are broadly removed after max-pooling. However, the coarse background image from max-pooling is rough in background areas,  causing false positives for the coarse cue. This accounts for the further performance enhancement with the fine cue, indicating that fine cue guided segmentation obtained by the coarse-to-fine background reconstruction is the most effective.

\begin{table}[t]
\centering
\caption{Quantitative cross-dataset evaluation of the proposed CrackCue built upon U-Net on retinal blood vessel segmentation with model trained on STARE~\cite{STARE}. 
}
\resizebox{1.0\linewidth}{!}{
\begin{tabular}{|c|c|c|c|c|}
\hline
\rule{0pt}{14pt} Datasets & Methods  & Acc & AUC & F1\\ \hline
\rule{0pt}{14pt}\multirow{2}{*}{{ DRIVE~\cite{DRIVE}}} & U-Net~\cite{UNET}  & 0.943 & 0.936 & 0.743\\ 
\rule{0pt}{12pt} & \cellcolor{gray!20}+CrackCue  &\cellcolor{gray!20} \textbf{0.947}  {\footnotesize (+0.4\%)}  &\cellcolor{gray!20} \textbf{0.940}  {\footnotesize (+0.4\%)} & \cellcolor{gray!20}\textbf{0.762} {\footnotesize (+1.9\%)} \\ \hline
\rule{0pt}{12pt}\multirow{2}{*}{{ CHASEDB1~\cite{CHASEDB1}}} & U-Net~\cite{UNET} & 0.930 & 0.908 & 0.603\\ 
\rule{0pt}{12pt} & \cellcolor{gray!20}+CrackCue  &\cellcolor{gray!20} \textbf{0.937}  {\footnotesize (+0.7\%)} & \cellcolor{gray!20}\textbf{0.934} {\footnotesize (+2.6\%)} &\cellcolor{gray!20} \textbf{0.645}  {\footnotesize (+4.2\%)} \\ \hline
\end{tabular}
}
\label{tab2}
\end{table}

\medskip
\noindent\textbf{CrackCue versus other variants of cue generation.} To confirm the benefits of the crack cues generated by the propoesed CrackCue, We employ median filtering, black hat transformation and the Canny edge detector, respectively to generate crack cues for assessment against those produced by CrackCue. The first type of cue is derived from the absolute difference between the median-filtered image and the input image, and the second from the absolute difference between the morphologically closed input image and the original image. The third type of cue is derived from the edge detection result of the Canny operator. The comparative experimental results are presented in Table~\ref{tab3}. The cues from the first two methods moderately improve network segmentation but are less effective than those from CrackCue. Additionally, cues generated by the Canny~\cite{canny1986computational} operator even leads to a decrease in detection performance. Their susceptibility to disturbances from complex backgrounds and shadows often results in numerous false positives, compromising their robustness.

\medskip
\noindent\textbf{Effect of using different dilated crack width $d$.} In Eq.~\eqref{final recon loss} for computing $L_{rec}$, the ground truth $Y$ is dilated to ensure $Y_d$ has a certain width. We conduct comparative experiments with different widths $d$. The experimental results are presented in Table~\ref{tab5}. The optimal performance is achieved when the width $d$ is set to 4 pixels.

\medskip
\noindent\textbf{Balancing weight $\lambda$ for two loss items in \eqref{total loss}.} We conduct ablation study on the balancing weight $\lambda$ for two loss items in \eqref{total loss}. The experimental results in Table~\ref{tab:lambda} demonstrate that CrackCue is approximately robust to $\lambda$, with the best results achieved when $\lambda$ is set to 1.

\medskip
\noindent\textbf{kernel size of max-pooling operator.} We conduct ablation study on the kernel size of the max-pooling operator used for eliminating cracks. The experimental results are shown in Table~\ref{tab:kernel}. When the kernel size is too small, the cracks cannot be completely eliminated. When the kernel size is too large, a substantial amount of image detail is lost, resulting in an overly coarse background image after upsampling. The best performance is achieved when the kernel size is set to 8.

\subsection{Runtime analysis}
We conduct the runtime analysis of three baseline networks and the corresponding segmentation networks guided by both coarse and fine cues generated by CrackCue. The results in Table \ref{runtime} indicate that guiding the network segmentation with coarse cue can effectively improve crack detection accuracy without impacting detection speed. It is noteworthy that generating coarse cue involves just a pair of max-pooling and upsampling operations, with negligible impact on inference time. Further generating fine cue, although leading to increased inference time, does not significantly affect the computational load for large networks like DeepCrack \cite{deepcrack} and CrackFormer \cite{liu2021crackformer}. Moreover, it can further optimize performance, making it an overall acceptable trade-off.

\subsection{Application to retinal blood vessel segmentation}
Since retinal vessels are also fine curvilinear objects, we conduct cross-dataset evaluations on retinal vessel datasets to verify whether CrackCue is equally effective for other curvilinear structures. The experimental setup is similar to that of the crack datasets, where the network is trained using the STARE~\cite{STARE} dataset and evaluated on the DRIVE~\cite{DRIVE} and CHASEDB1~\cite{CHASEDB1} datasets. We evaluate performance using the classical metrics: Accuracy (Acc), Area Under the Receiver Operating Characteristic Curve (AUC) and the F1 score. The baseline network is U-Net~\cite{UNET}, compared against U-Net enhanced with CrackCue. The quantitative results of the retinal cross-dataset evaluation are shown in Table~\ref{tab2}. Incorporating CrackCue improves or matches baseline metrics across all tests. Notably, in the STARE to CHASEDB1 generalization experiment,  the performance of U-Net is not ideal. However, after integrating CrackCue, the F1 score increases by 4.2\%, significantly improving the performance. The results of the aforementioned experiments illustrate that the proposed CrackCue effectively enhances the robustness of curvilinear object segmentation networks, proving its applicability in segmenting various types of curvilinear objects.

\subsection{Limitation}
\label{subsec:limitation}
The proposed CrackCue is based on the property that cracks are usually thin structures, which is true in most cases. One potential limitation of the proposed method is that the proposed CrackCue may be less effective in detecting very wide cracks. This is because downsampling operation may not fully eliminate wide cracks, failing to generate appropriate crack cues. Yet, such wide cracks are usually quite obvious and easy to detect.

\section{Conclusion}
\label{sec:blind}
In this paper, we aim to address the issue of unsatisfied generalization ability to unseen domains for existing crack detection methods. To this end, we introduce a novel plug-and-play method termed CrackCue. By reconstructing the crack-free background image through a coarse-to-fine reconstruction network, our method provides robust crack cues about the approximate location of cracks to the segmentation network. Experiments on four widely-used crack datasets demonstrate that the proposed CrackCue significantly improves cross-dataset performance and enhances the robustness of the crack detection network. Furthermore, extension experiments on retinal blood vessel segmentation also show that our CrackCue is not limited to crack detection, but can be extended to other tubular structure image segmentation tasks. In the future, we would like to explore the combination of our CrackCue with other domain generalization methods to further boost the generalizability of crack detection methods.

\section*{Acknowledgement}
    This work was supported partially by the National Key Research and Development Program of China (2023YFC2705700), the National Natural Science Foundation of China (62222112 and 62176186).
	\bibliographystyle{unsrt} 
	\bibliography{main}

\begin{thebibliography}{10}

\bibitem{chen2024mind}
Zhuangzhuang Chen, Zhuonan Lai, Jie Chen, and Jianqiang Li.
\newblock Mind marginal non-crack regions: Clustering-inspired representation
  learning for crack segmentation.
\newblock In {\em Proc. of IEEE Conf. on Computer Vision and Pattern
  Recognition}, pages 12698--12708, 2024.

\bibitem{zou2012cracktree}
Qin Zou, Yu~Cao, Qingquan Li, Qingzhou Mao, and Song Wang.
\newblock Cracktree: Automatic crack detection from pavement images.
\newblock {\em Pattern Recognition Letters}, 33(3):227--238, 2012.

\bibitem{Crackforest}
Yong Shi, Limeng Cui, Zhiquan Qi, Fan Meng, and Zhensong Chen.
\newblock Automatic road crack detection using random structured forests.
\newblock {\em IEEE Trans. on Intelligent Transportation Systems},
  17(12):3434--3445, 2016.

\bibitem{HED}
Saining Xie and Zhuowen Tu.
\newblock Holistically-nested edge detection.
\newblock In {\em Proc. of IEEE Intl. Conf. on Computer Vision}, pages
  1395--1403, 2015.

\bibitem{rcf}
Yun Liu, Ming-Ming Cheng, Xiaowei Hu, Kai Wang, and Xiang Bai.
\newblock Richer convolutional features for edge detection.
\newblock In {\em Proc. of IEEE Conf. on Computer Vision and Pattern
  Recognition}, pages 3000--3009, 2017.

\bibitem{deepcrack}
Qin Zou, Zheng Zhang, Qingquan Li, Xianbiao Qi, Qian Wang, and Song Wang.
\newblock Deepcrack: Learning hierarchical convolutional features for crack
  detection.
\newblock {\em IEEE Trans. on Image Processing}, 28(3):1498--1512, 2018.

\bibitem{quan2023crackvit}
Jianing Quan, Baozhen Ge, and Min Wang.
\newblock Crackvit: a unified cnn-transformer model for pixel-level crack
  extraction.
\newblock {\em Neural Computing and Applications}, pages 1--17, 2023.

\bibitem{liu2021crackformer}
Huajun Liu, Xiangyu Miao, Christoph Mertz, Chengzhong Xu, and Hui Kong.
\newblock Crackformer: Transformer network for fine-grained crack detection.
\newblock In {\em Proc. of IEEE Intl. Conf. on Computer Vision}, pages
  3783--3792, 2021.

\bibitem{tang201}
Jinshan Tang and Yanliang Gu.
\newblock Automatic crack detection and segmentation using a hybrid algorithm
  for road distress analysis.
\newblock In {\em 2013 IEEE International Conference on Systems, Nan, and
  Cybernetics}, pages 3026--3030, 2013.

\bibitem{yu2019dark}
Li~Yu, Yugang Tian, and Wei Wu.
\newblock A dark target detection method based on the adjacency effect: A case
  study on crack detection.
\newblock {\em Sensors}, 19(12):2829, 2019.

\bibitem{canny1986computational}
John Canny.
\newblock A computational approach to edge detection.
\newblock {\em IEEE Trans. on Pattern Anal. and Mach. Intell.}, pages 679--698,
  1986.

\bibitem{tabatabaei2019automatic}
Seyed Amir~Hossein Tabatabaei, Ahmad Delforouzi, Muhammad~Hassan Khan, Tim
  Wesener, and Marcin Grzegorzek.
\newblock Automatic detection of the cracks on the concrete railway sleepers.
\newblock {\em International Journal of Pattern Recognition and Artificial
  Intelligence}, 33(09):1955010, 2019.

\bibitem{adaboost}
Aur{\'e}lien Cord and Sylvie Chambon.
\newblock Automatic road defect detection by textural pattern recognition based
  on adaboost.
\newblock {\em Computer-Aided Civil and Infrastructure Engineering},
  27(4):244--259, 2012.

\bibitem{UNET}
Olaf Ronneberger, Philipp Fischer, and Thomas Brox.
\newblock U-net: Convolutional networks for biomedical image segmentation.
\newblock In {\em Proc. of Intl. Conf. on Medical Image Computing and Computer
  Assisted Intervention}, pages 234--241, 2015.

\bibitem{segnet}
Vijay Badrinarayanan, Alex Kendall, and Roberto Cipolla.
\newblock Segnet: A deep convolutional encoder-decoder architecture for image
  segmentation.
\newblock {\em IEEE Trans. on Pattern Anal. and Mach. Intell.},
  39(12):2481--2495, 2017.

\bibitem{barisin2022methods}
Tin Barisin, Christian Jung, Franziska M{\"u}sebeck, Claudia Redenbach, and
  Katja Schladitz.
\newblock Methods for segmenting cracks in 3d images of concrete: A comparison
  based on semi-synthetic images.
\newblock {\em Pattern Recognition}, 129:108747, 2022.

\bibitem{fang2020novel}
Fen Fang, Liyuan Li, Ying Gu, Hongyuan Zhu, and Joo-Hwee Lim.
\newblock A novel hybrid approach for crack detection.
\newblock {\em Pattern Recognition}, 107:107474, 2020.

\bibitem{mei2020densely}
Qipei Mei, Mustafa G{\"u}l, and Md~Riasat Azim.
\newblock Densely connected deep neural network considering connectivity of
  pixels for automatic crack detection.
\newblock {\em Automation in Construction}, 110:103018, 2020.

\bibitem{FPHBN}
Fan Yang, Lei Zhang, Sijia Yu, Danil Prokhorov, Xue Mei, and Haibin Ling.
\newblock Feature pyramid and hierarchical boosting network for pavement crack
  detection.
\newblock {\em IEEE Trans. on Intelligent Transportation Systems},
  21(4):1525--1535, 2019.

\bibitem{CASA-Net}
Xin Bi, Shining Zhang, Yu~Zhang, Lei Hu, Wei Zhang, Wenjing Niu, Ye~Yuan, and
  Guoren Wang.
\newblock Casa-net: a context-aware correlation convolutional network for
  scale-adaptive crack detection.
\newblock In {\em Proceedings of the 31st ACM International Conference on
  Information \& Knowledge Management}, pages 67--76, 2022.

\bibitem{CrackDet}
Zhuangzhuang Chen, Jin Zhang, Zhuonan Lai, Guanming Zhu, Zun Liu, Jie Chen, and
  Jianqiang Li.
\newblock The devil is in the crack orientation: A new perspective for crack
  detection.
\newblock In {\em Proc. of IEEE Intl. Conf. on Computer Vision}, pages
  6653--6663, 2023.

\bibitem{bang2019encoder}
Seongdeok Bang, Somin Park, Hongjo Kim, and Hyoungkwan Kim.
\newblock Encoder--decoder network for pixel-level road crack detection in
  black-box images.
\newblock {\em Computer-Aided Civil and Infrastructure Engineering},
  34(8):713--727, 2019.

\bibitem{park2019patch}
Somin Park, Seongdeok Bang, Hongjo Kim, and Hyoungkwan Kim.
\newblock Patch-based crack detection in black box images using convolutional
  neural networks.
\newblock {\em Journal of Computing in Civil Engineering}, 33(3):04019017,
  2019.

\bibitem{lu2025deep}
Xiaochun Lu, Qingquan Li, Jianyuan Li, and La~Zhang.
\newblock Deep learning-based method for detection and feature quantification
  of microscopic cracks on the surface of concrete dams.
\newblock {\em Measurement}, 240:115587, 2025.

\bibitem{tong2025modulated}
Shuzhen Tong, Qing Wang, Xuan Wei, Cheng Lu, and Xiaobo Lu.
\newblock Modulated deformable convolution based on graph convolution network
  for rail surface crack detection.
\newblock {\em Signal Processing: Image Communication}, 130:117202, 2025.

\bibitem{chen2022refined}
Junzhou Chen, Nan Zhao, Ronghui Zhang, Long Chen, Kai Huang, and Zhijun Qiu.
\newblock Refined crack detection via lecsformer for autonomous road inspection
  vehicles.
\newblock {\em IEEE Transactions on Intelligent Vehicles}, 8(3):2049--2061,
  2022.

\bibitem{kuang2024universal}
Senyun Kuang, Yang Liu, Xin Wang, Xiaobo Qu, and Yintao Wei.
\newblock An universal crack detection framework for intelligent
  road-perceptive vehicles.
\newblock {\em IEEE Transactions on Intelligent Vehicles}, 2024.

\bibitem{ma2024transformer}
Mingyang Ma, Lei Yang, Yanhong Liu, and Hongnian Yu.
\newblock A transformer-based network with feature complementary fusion for
  crack defect detection.
\newblock {\em IEEE Trans. on Intelligent Transportation Systems}, 2024.

\bibitem{zhao2024crack}
Yuanlin Zhao, Wei Li, Jiangang Ding, Yansong Wang, Lili Pei, and Aojia Tian.
\newblock Crack instance segmentation using splittable transformer and position
  coordinates.
\newblock {\em Automation in Construction}, 168:105838, 2024.

\bibitem{yu2024robust}
Zhenwei Yu, Qinyu Chen, Yonggang Shen, and Yiping Zhang.
\newblock Robust pavement crack segmentation network based on transformer and
  dual-branch decoder.
\newblock {\em Construction and Building Materials}, 453:139026, 2024.

\bibitem{hu2025ccdformer}
Xiangkun Hu, Hua Li, Yixiong Feng, Songrong Qian, Jian Li, and Shaobo Li.
\newblock Ccdformer: A dual-backbone complex crack detection network with
  transformer.
\newblock {\em Pattern Recognition}, 161:111251, 2025.

\bibitem{shan2025dcuformer}
Jinhuan Shan, Yue Huang, and Wei Jiang.
\newblock Dcuformer: Enhancing pavement crack segmentation in complex
  environments with dual-cross/upsampling attention.
\newblock {\em Expert Systems with Applications}, 264:125891, 2025.

\bibitem{liu2025scsegamba}
Hui Liu, Chen Jia, Fan Shi, Xu~Cheng, and Shengyong Chen.
\newblock Scsegamba: Lightweight structure-aware vision mamba for crack
  segmentation in structures.
\newblock In {\em Proc. of IEEE Conf. on Computer Vision and Pattern
  Recognition}, 2025.

\bibitem{zhang2025crack}
Jianming Zhang, Shigen Zhang, Dianwen Li, Jianxin Wang, and Jin Wang.
\newblock Crack segmentation network via difference convolution-based encoder
  and hybrid cnn-mamba multi-scale attention.
\newblock {\em Pattern Recognition}, page 111723, 2025.

\bibitem{liot}
Tianyi Shi, Nicolas Boutry, Yongchao Xu, and Thierry G{\'e}raud.
\newblock Local intensity order transformation for robust curvilinear object
  segmentation.
\newblock {\em IEEE Trans. on Image Processing}, 31:2557--2569, 2022.

\bibitem{hendrycks2019benchmarking}
Dan Hendrycks and Thomas Dietterich.
\newblock Benchmarking neural network robustness to common corruptions and
  perturbations.
\newblock {\em Proc. of International Conference on Learning Representations},
  2018.

\bibitem{liu2023crackformer}
Huajun Liu, Jing Yang, Xiangyu Miao, Christoph Mertz, and Hui Kong.
\newblock Crackformer network for pavement crack segmentation.
\newblock {\em IEEE Trans. on Intelligent Transportation Systems},
  24(9):9240--9252, 2023.

\bibitem{STARE}
AD~Hoover, Valentina Kouznetsova, and Michael Goldbaum.
\newblock Locating blood vessels in retinal images by piecewise threshold
  probing of a matched filter response.
\newblock {\em IEEE Trans. on Medical Imaging}, 19(3):203--210, 2000.

\bibitem{DRIVE}
Joes Staal, Michael~D Abr{\`a}moff, Meindert Niemeijer, Max~A Viergever, and
  Bram Van~Ginneken.
\newblock Ridge-based vessel segmentation in color images of the retina.
\newblock {\em IEEE Trans. on Medical Imaging}, 23(4):501--509, 2004.

\bibitem{CHASEDB1}
Muhammad~Moazam Fraz, Paolo Remagnino, Andreas Hoppe, Bunyarit Uyyanonvara,
  Alicja~R Rudnicka, Christopher~G Owen, and Sarah~A Barman.
\newblock An ensemble classification-based approach applied to retinal blood
  vessel segmentation.
\newblock {\em IEEE Transactions on Biomedical Engineering}, 59(9):2538--2548,
  2012.

\end{thebibliography}

\end{document}